\documentclass[letterpaper]{article} 
\usepackage{aaai2026}  
\usepackage{times}  
\usepackage{helvet}  
\usepackage{courier}  
\usepackage[hyphens]{url}  
\usepackage{graphicx} 
\usepackage{amsmath}
\usepackage{amsfonts}
\usepackage{cleveref}
\usepackage{amssymb, pifont}
\usepackage{tabularx}
\usepackage{booktabs}
\usepackage{enumitem}
\usepackage{natbib}  
\usepackage{caption} 
\usepackage{algorithm}
\usepackage{algorithmic}

\usepackage{newfloat}
\usepackage{listings}
\DeclareCaptionStyle{ruled}{labelfont=normalfont,labelsep=colon,strut=off} 
\floatstyle{ruled}
\newfloat{listing}{tb}{lst}{}
\floatname{listing}{Listing}
\title{Conditional Information Bottleneck for Multimodal Fusion: Overcoming Shortcut Learning in Sarcasm Detection}

\author {
    Yihua Wang\textsuperscript{\rm 1, 2}, 
    Qi Jia\textsuperscript{\rm 2}, 
    Cong Xu\textsuperscript{\rm 2}\thanks{Corresponding Authors.\\ Work performed during an internship at IEIT SYSTEMS Co., Ltd.}, 
    Feiyu Chen\textsuperscript{\rm 1}, 
    Yuhan Liu\textsuperscript{\rm 1}, 
    Haotian Zhang\textsuperscript{\rm 1}, 
    Liang Jin\textsuperscript{\rm 2}, 
    Lu Liu\textsuperscript{\rm 2}, 
    Zhichun Wang\textsuperscript{\rm 1}\footnotemark[1]
}
\affiliations {
     \textsuperscript{\rm 1}School of Artificial Intelligence, Beijing Normal University, Beijing, China\\
    \textsuperscript{\rm 2}IEIT SYSTEMS Co., Ltd., Beijing, China\\
    wangyihua@mail.bnu.edu.cn, xucong@ieisystem.com, zcwang@bnu.edu.cn
}

\usepackage{bibentry}

\begin{document}

\maketitle

\begin{abstract}
Multimodal sarcasm detection is a complex task that requires distinguishing subtle complementary signals across modalities while filtering out irrelevant information. Many advanced methods rely on learning shortcuts from datasets rather than extracting intended sarcasm-related features. However, our experiments show that shortcut learning impairs the model's generalization in real-world scenarios. Furthermore, we reveal the weaknesses of current modality fusion strategies for multimodal sarcasm detection through systematic experiments, highlighting the necessity of focusing on effective modality fusion for complex emotion recognition. To address these challenges, we construct MUStARD++$^{R}$ by removing shortcut signals from MUStARD++. Then, a Multimodal Conditional Information Bottleneck (MCIB) model is introduced to enable efficient multimodal fusion for sarcasm detection. Experimental results show that the MCIB achieves the best performance without relying on shortcut learning.
\end{abstract}

\begin{links}
    \link{Code}{https://github.com/sljgkjhwe/MCIB.git}
    \link{Datasets}{https://pan.quark.cn/s/975e0d976744#/list/share} (3BFf)
\end{links}

\section{Introduction}

\begin{figure}[t]
  \centering
  \includegraphics[width=0.9\linewidth]{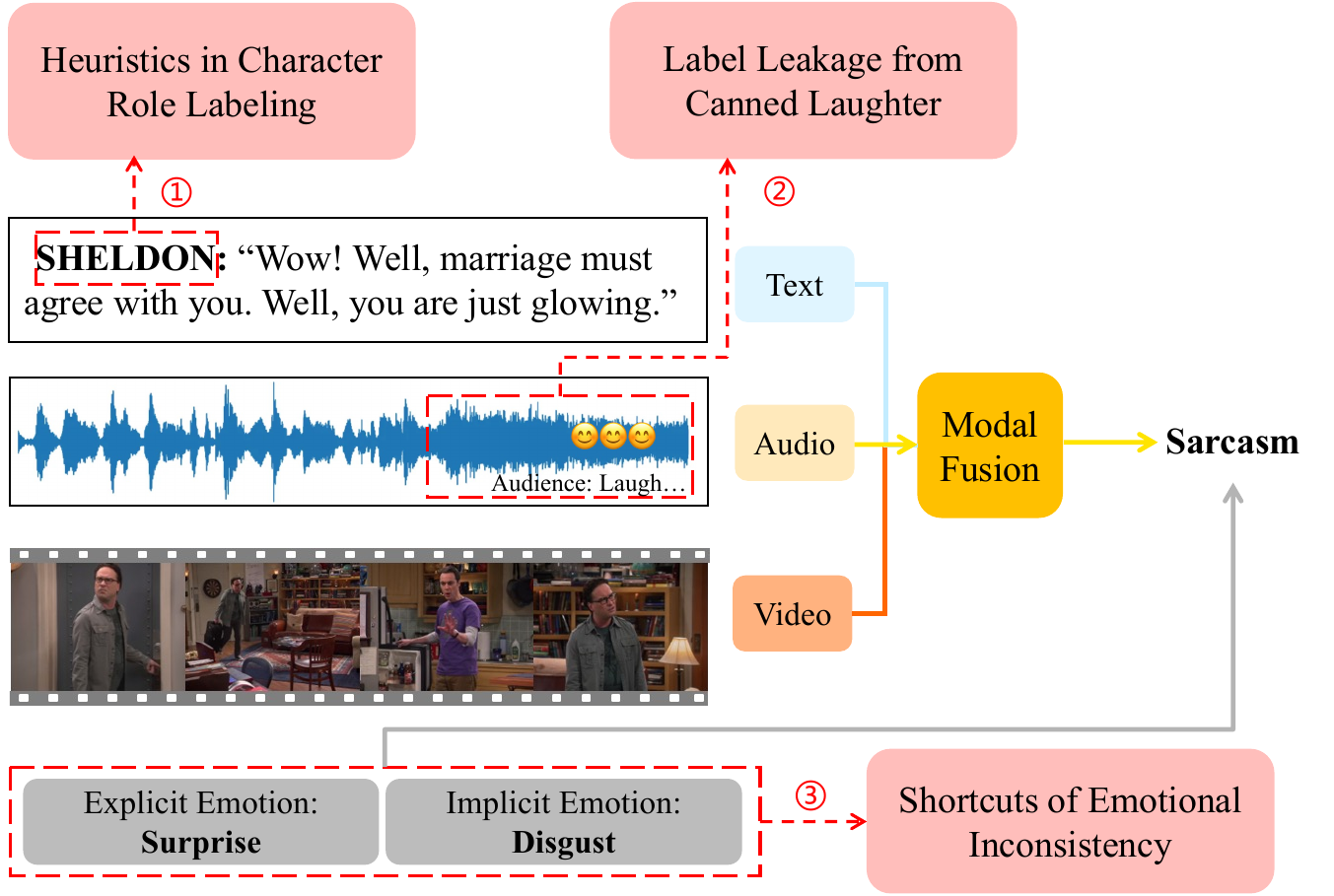}
  \caption{Multimodal sarcasm analysis often relies on sitcoms like Friends, which primarily feature character dialogues. The illustration presents the multimodal sarcasm detection task and highlights several shortcut learning issues.}
  \label{fig:fig1}
\end{figure}

Multimodal sentiment analysis (MSA) \cite{zadeh2017tensor} integrates various modalities to interpret human emotions, sentiments, and opinions. In MSA tasks, multimodal sarcasm detection \cite{castro2019towards} is particularly challenging due to the subtle contrast between surface meaning and underlying intent, which may convey humor, ridicule, or contempt \cite{joshi2017automatic}. See Figure \ref{fig:fig1}, Sheldon’s words express surprise, but his facial expression shows disgust, and his tone is mocking, conveying sarcasm. Despite advances in large language models, their performance remains capped by supervised baselines, falling short of human-level understanding \cite{hu2025emobenchmbench}.

Numerous studies have explored multimodal sarcasm detection (MSD) from various perspectives. Many focus on the inherent characteristics of sarcasm or are specifically designed for prior information: such as contrasting emotional polarity \cite{lu2024fact, wang2024cross}, or incorporate external knowledge \cite{yue2023knowlenet}, character profiles \cite{li2024attention}, emotional cues \cite{gao2024improving}, and fine-grained visual features \cite{tiwari2023predict}.

As shown by issues 1, 2, and 3 in Figure \ref{fig:fig1}, models tend to learn specific shortcuts. Those shortcuts \cite{geirhos2020shortcut} often fail to generalize to more diverse conditions, typically in real-world scenarios. We identify three major shortcut issues in multimodal sarcasm detection research with the representative dataset MUStARD++ \cite{ray2022multimodal}. (1) Heuristics in Character Role Labeling: In sarcasm datasets, some characters exhibit a speaking style that favors sarcasm, such as Sheldon in the TV show THE BIG BANG THEORY. Introducing these character labels during training causes the model to form a bias toward sarcasm associated with a specific person. (2) Label Leakage from Canned Laughter: Canned laughter (artificial background laughter) often causes label leakage. A review of MUStARD++ reveals that canned laughter frequently follows sarcastic utterances, but rarely non-sarcastic ones. Thus, the model tends to classify samples with canned laughter as sarcasm. (3) Shortcuts of Emotional Inconsistency: The dataset provides implicit/explicit emotion labels, and their contrast serves as a strong indicator of sarcasm. Some models are designed to learn implicit/explicit emotions rather than true sarcasm, limiting their ability to detect sarcasm due to over-reliance on inconsistency.

The key to understanding the intended complex information and genuinely improving performance is addressing the core multimodal fusion problem, rather than designing modules based on shortcut learning. Various fusion methods \cite{ray2022multimodal, zhang2024self,zhang2023learning} do not result in significant information gains, some even reducing task performance, highlighting their poor generalization and adaptability to sarcasm tasks. Higher modal complementarity leads to poorer robustness when certain modes are absent or misaligned, and noise has a greater impact on modes with high complementarity \cite{li2023modality}. Multimodal sarcasm detection involves heterogeneous information with complementary but also irrelevant or misleading redundancy. Therefore, this underscores the need for an efficient fusion strategy that integrates complementary information while mitigating redundancy.

To this end, we construct MUStARD++$^{R}$ by removing canned laughter, character labels, and emotional polarity from the dataset MUStARD++.
Meanwhile, we propose a robust \textbf{M}ultimodal fusion framework with a \textbf{C}onditional \textbf{I}nformation \textbf{B}ottleneck for sarcasm detection, named \textbf{MCIB}. The MCIB achieves the state-of-the-art performance solely through effective multimodal fusion, without relying on additional shortcut cues. The MCIB offers two key functionalities. First, it addresses the limitations of traditional information bottlenecks by enabling the fusion of more than two information sources, overcoming the challenge of processing multiple modalities simultaneously. Second, our multimodal fusion strategy is designed to extract and separate inconsistent yet relevant information from each modality, effectively reducing inter-modal redundancy and preserving critical features for more accurate multimodal sarcasm recognition.

Our main contributions: (1) We conduct an in-depth analysis and provide a comprehensive summary of the critical issues in multimodal sarcasm detection, addressing the issue of learning shortcuts and refactoring the benchmark. (2) We rethink multimodal fusion from a novel perspective and propose a flexible and robust multimodal conditional information bottleneck fusion method. (3) We perform extensive experiments to demonstrate the performance of our fusion method, showing that the model significantly improves the efficiency of extracting meaningful information that aids in sarcasm detection across multiple modalities.

\section{Related Work}

\textbf{Multimodal Sarcasm Detection} has evolved from text and image-text \cite{wen2023dip} to video clips, which better reflect real-life scenarios. Many methods leverage sarcasm traits to aid recognition, such as incongruity \cite{lu2024fact}. Song et al. \cite{song2024utterance} propose an utterance-level attention and incongruity learning network to capture incongruity representations in sarcastic expressions. Since sarcasm is closely linked to emotions and sentiments, an adaptive representation learning model \cite{zhang2024self} is proposed based on an expert network to analyze the emotion and sarcasm tasks jointly. Chauhan et al. \cite{chauhan2020sentiment} design a multitasking collaborative framework for cross-training and sharing attention between sarcasm, implicit/explicit emotions, and implicit/explicit sentiments. Additionally, Zhang et al. \cite{zhang2023learning} employ a dual-gating network and three separate layers to achieve cross-modal interactive learning. Other research attempts to leverage various external information. Studies such as \cite{liang2022multi} and \cite{yue2023knowlenet} have utilized tools like KnowleNet to implement knowledge fusion networks for MSD. Moreover, works like \cite{tomar2024action}, \cite{tiwari2023predict}, and \cite{pandey2025vyang} enhance sarcasm recognition by incorporating conversational behavioral cues, fine-grained eye gaze information, or deep visual features acquired through self-conditioning ConvNet, respectively. While \cite{gao2024improving, helal2024contextual} rely on pre-training with explicit and implicit emotions, the best results from \cite{li2024attention} similarly through character labeling. Besides, some novel approaches \cite{liu2023quantum, tiwari2024quantum} suggest that quantum mechanics are well-suited for capturing the complexity and uncertainty in sarcasm through a quantum probability-driven model or quantum fuzzy neural networks.

\textbf{Information Bottleneck} (IB) based approaches \cite{slonim1999agglomerative, alemi2022deep} have demonstrated their effectiveness in various tasks such as cross-modal clustering \cite{yan2023cross} and representation learning \cite{zhang2022unimodal, ding2023robust}, and have also been widely applied in multimodal fusion. For example, Zhang et al. \cite{zhang2022information} improve multimodal sentiment analysis performance by applying IB constraints to pairs of modalities. Mai et al. \cite{mai2022multimodal} ensure that latent modal representations can effectively handle the target task by introducing IB constraints between each modality and the predicted target. Chen et al. \cite{chen2023multimodal} adopt a similar approach, enhancing multimodal fusion by adding cross-attention. Furthermore, Liu et al. \cite{liu2024multimodal} apply IB during the representation fusion stage to discard irrelevant information from individual modalities. Xiao et al. \cite{xiao2024neuro} propose a two-layer IB structure that minimizes mutual information between input and latent states while maximizing it between latent and residual states. Moreover, conditional mutual information \cite{wyner1978definition} has demonstrated strong generalization performance in various applications, such as feature selection \cite{fleuret2004fast}, modality enhancement \cite{ji2022increasing}, and multimodal selection \cite{he2024efficient}. Gondek et al. \cite{gondek2003conditional} propose maximizing conditional mutual information to obtain relevant but novel clustering solutions, avoiding redundant information already captured by known structures or categories. Li et al. \cite{li2023modality} demonstrate through CIB calculations that higher modal complementarity reduces robustness while noise impacts highly complementary modes more severely. Thus, recognizing the potential of CIB \cite{molavipour2021neural} for extracting relevant information in multimodal scenarios, we design a CIB-based method for multimodal fusion.

\section{Task Analysis and Refactoring}
\label{sec:motiv}
In the MSD task, some shortcuts arise from task-specific traits, while others capture superficial cues or spurious correlations rather than the true intent of sarcasm. These easy-to-learn shortcut signals hinder the model's generalization to broader testing environments.

\textbf{Heuristics in Character Role Labeling.} Capturing character-specific traits can improve sarcasm detection, but methods relying on character association lack generalizability, as such features are often unavailable in real-world settings. We analyzed the proportion of sarcasm and non-sarcasm in the dialogues of different characters in the MUStARD++. The chi-square test is used to examine the relationship between character and sarcasm, with the null hypothesis stating no association. The test statistic of 166.7 and p-value of $3.89 \times 10^{-27}$ provide strong evidence against the null hypothesis. The actors exhibited a distinct tendency towards either sarcastic or non-sarcastic behavior, suggesting that character traits strongly influence the judgment of sarcasm. The experimental findings \cite{castro2019towards, li2024attention} reveal that the majority of models can enhance sarcasm recognition merely by incorporating character embeddings. However, the performance of such models trained with character embeddings will drop drastically when tested on other sentiment tasks. Thus, it is essential to design a robust model that captures the character's intended features rather than relying on character labels.


\textbf{Label Leakage from Canned Laughter.} Since most sarcasm datasets are derived from sitcoms, they typically include canned laughter. Canned laughter, a pre-recorded sound used to enhance humor or guide audience reactions, is also prevalent in MUStARD++. It often follows ironic or humorous utterances, leading to label leakage. Most studies \cite{arora2023universlu, tomar2023your, li2024attention} ignore the canned laughter while processing the data. Using the pre-trained speech recognition model SpeechPrompt v2\cite{chang2023speechprompt}, we validated on MUStARD++ and MUStARD++$^{R}$ (with and without canned laughter). Compared to MUStARD++, performance on MUStARD++$^{R}$ shows a significant drop, with F1-score decreasing from 73.47 to 43.59 (a drop of 29.88) and accuracy from 78.33 to 63.03 (a drop of 15.3). This suggests that the model learned the features associated with canned laughter, indirectly confirming the label leakage issue.

\textbf{Shortcuts of Emotional Inconsistency.} Sarcasm is often a sugar-coated bomb, expressing the opposite of its literal meaning by masking implicit emotions with explicit ones. The statistical results from the MUStARD++ dataset reveal that sarcasm is predominantly associated with different emotions, accounting for $99\%$ of cases, while only $1\%$ is linked to the same emotions. In contrast, non-sarcasm is mainly associated with the same emotions ($94.7\%$), with only $5.3\%$ corresponding to different emotions. This indicates that sarcastic sentences often exhibit discrepancies between explicit and implicit sentiment labels. The Phi coefficient analysis supports this observation, revealing a very strong association between emotional consistency and sarcasm ($\phi = 0.94$). Many studies \cite{chauhan2020sentiment, shah2022emotion, helal2024contextual} use this pattern to detect sarcasm. 
But in real-world scenarios, the explicit/implicit emotion labels do not exist. Moreover, inferring explicit/implicit emotion labels may be more challenging than directly detecting sarcasm. 
These limitations suggest that relying on explicit/implicit emotion labels for MSD is unfeasible.

To avoid relying on shortcuts and improve the fairness, robustness, trustworthiness, and deployability of the evaluation benchmark, based on the above analysis of shortcut learning in multimodal sarcasm tasks, we reconstructed MUStARD++$^{R}$ from the publicly available dataset MUStARD++. Each data sample is annotated with sarcasm labels, implicit and explicit emotions, arousal, and valence tags. The first step in modifying the dataset is to remove all shortcut labels, leaving only those related to sarcasm. Next, we eliminated video segments that contain canned laughter. Specifically, we used the timestamp of the first word in each utterance as the start of the video segment and the timestamp of the last word as the endpoint. By removing labels linked to shortcut learning, MUStARD++$^{R}$ allows for a more accurate evaluation of multimodal fusion models in detecting sarcasm through integrated cross-modal information.

\section{Methodology}

This section defines the formal optimization objective for multimodal fusion, rethinks existing fusion methods, and introduces our framework, including the MCIB algorithm and overall model architecture.


\noindent \textbf{Notations.} We define the modalities audio, video, and text as $\{x_0, x_1, x_2\}$, which are interdependent random variables. The redundant information causing interference among modalities is denoted by $R$, while the information gain from modality complementarity is represented by $C$. $y$ denotes the ground truth of the target task.

\begin{figure*}[t]
  \centering
  \includegraphics[width=0.9\linewidth]{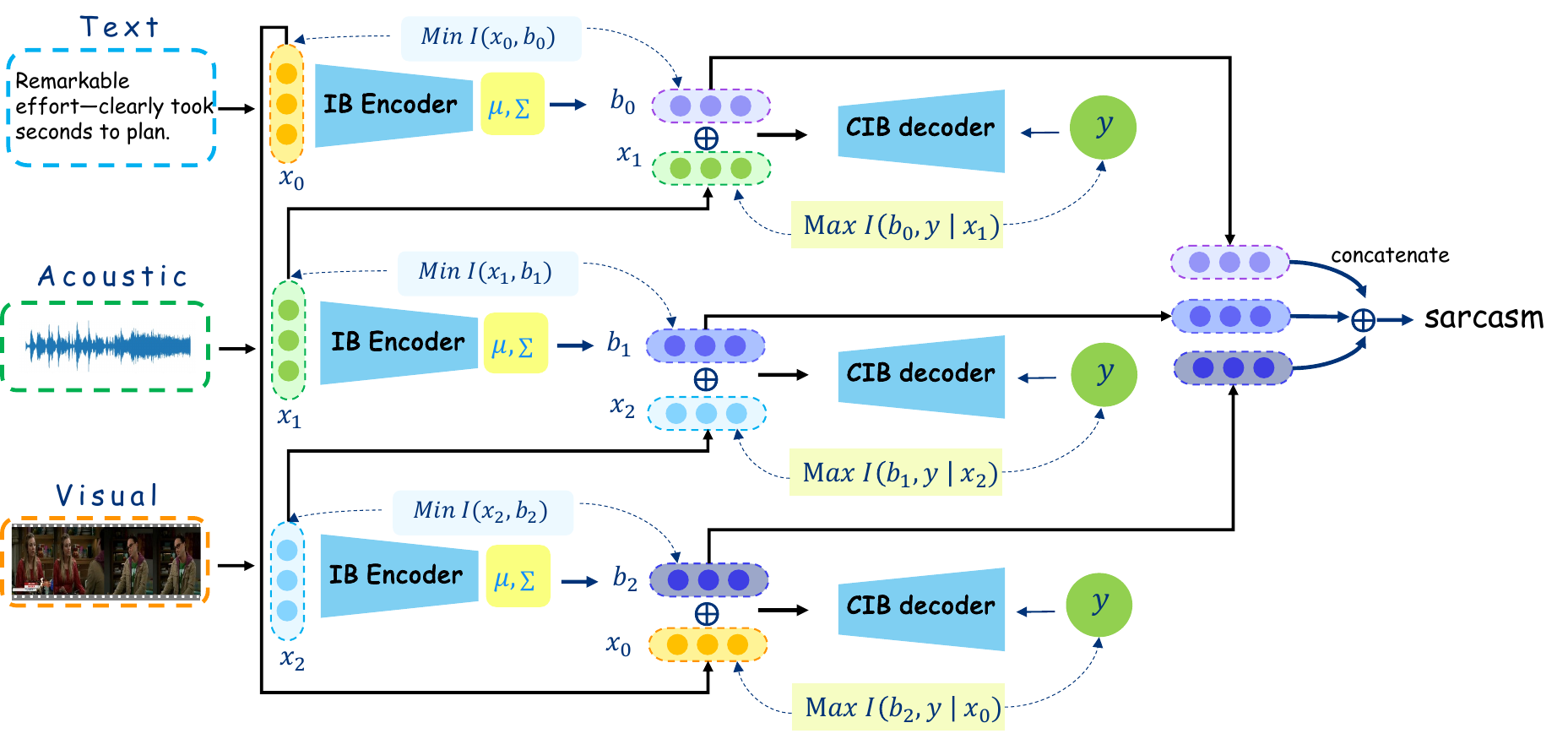}
  \caption{The diagram illustrates the overall architecture of the MCIB model. The multimodal fusion component employs three parallel conditional information bottleneck structures to filter out irrelevant information and extract relevant information between each pair of modalities. For each pair of modalities, we first minimize the mutual information between the primary modality and the latent state to achieve filtering and compression through the information bottleneck. We then maximize the conditional mutual information among the auxiliary modality, latent state, and prediction target. Finally, the bidirectional optimization within CIB produces an intermediate representation $b$ that encapsulates the essential information required for our prediction target.}
  \label{fig:min}
\end{figure*}

\noindent \textbf{Problem formulation.}  
Our objective is to minimize the loss between the fused multimodal information and the target $y$. As shown in \cref{eq:1}, we aim to reduce redundancy $R$ and maximize the utilization of complementarity $C$ between modalities:
\begin{equation}
    \begin{aligned}
        \min_{Fusion} \quad & Loss\left( y, f(x_0, x_1, x_2) \right), \\
        \text{subject to} \quad & R(x_0, x_1, x_2; y) \leq \delta \\
        & C(x_0, x_1, x_2; y) \geq \epsilon
    \end{aligned}
    \label{eq:1}
\end{equation}
where $\delta$ sets the upper limit for allowable redundancy, and $\epsilon$ defines the minimum threshold for the required level of complementarity utilization among modalities.

\begin{figure}[H]
  \centering
  \includegraphics[width=0.9\linewidth]{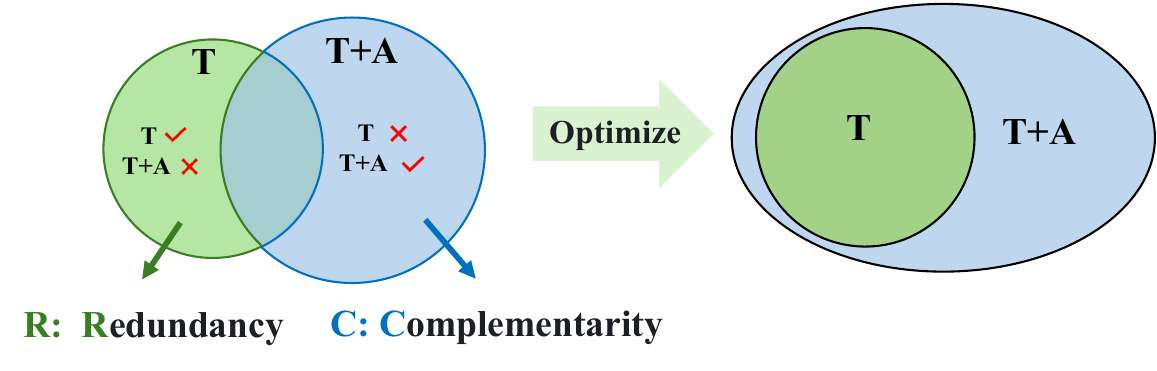}
  \caption{The diagram illustrates the optimization direction for multimodal fusion. The two circles represent correctly predicted samples by the Text modality alone and by the Text + Acoustic combination. The overlapping region shows samples correctly predicted by both T and T + A; region R represents samples correctly predicted by T but misclassified when A is added; and region C represents samples only correctly predicted with T + A, while T alone fails.}
  \label{fig:mxr}
\end{figure}

\noindent \textbf{Modal Fusion Rethink.} We rethink the limited effectiveness of previous fusion methods in light of the complementary and redundant properties of multimodal data. Ablation tests on various combinations of modality reveal limited gains from modality fusion. In some cases, adding modalities reduces performance: audio may decrease overall effectiveness. Figure \ref{fig:mxr} illustrates the reasons for ineffective modality fusion: the Redundancy region (which refers not only to "repeated information" that may be useless for the object but also to negative interference in modality fusion) is larger than the Complementarity region, and the overlapping middle region is too small. This suggests that misleading redundant information outweighs the informational gain from the added modality. For effective modality addition, irrelevant information that may mislead predictions should be reduced, while enhancing the benefits of complementary useful information.

\subsection{Our Framework}
To extract complementary information and remove redundancy across modalities, we apply the CIB principle to build a multimodal fusion model (see Fig.~\ref{fig:min}). We first extract detailed modality features, then design a fusion algorithm based on multimodal pairwise strategy. Compact cross-modal representations are optimized to enhance target understanding, and a balanced, robust loss function with Lagrange constraints ensures effective model training.

\subsubsection{MCIB Algorithm}
To achieve the multimodal fusion objective in \cref{eq:1}, we distinguish between the primary modality $x_p$ and the auxiliary modality $x_a$, where $x_p, x_a \in \{x_0, x_1, x_2\}$. Only the information in the primary modality $x_p$ beneficial to the target is preserved, while the auxiliary modality $x_a$ provides complementary information to $x_p$. We aim to encode as much task-related ($y$) information as possible in the latent state $b$, where $b \in \{b_0, b_1, b_2\}$ represents generation through optimization of the conditional information bottleneck. As shown in Figure \ref{fig:mm}, the trade-off between compressing redundant information from the primary modality and retaining complementary information provided by the auxiliary modality is optimized by minimizing mutual information while adhering to specific constraints. We reformulate this constrained optimization problem into an unconstrained Lagrangian form \cref{eq:2}, known as the conditional information bottleneck:

\begin{figure}[h]
  \centering
  \includegraphics[width=0.8\linewidth]{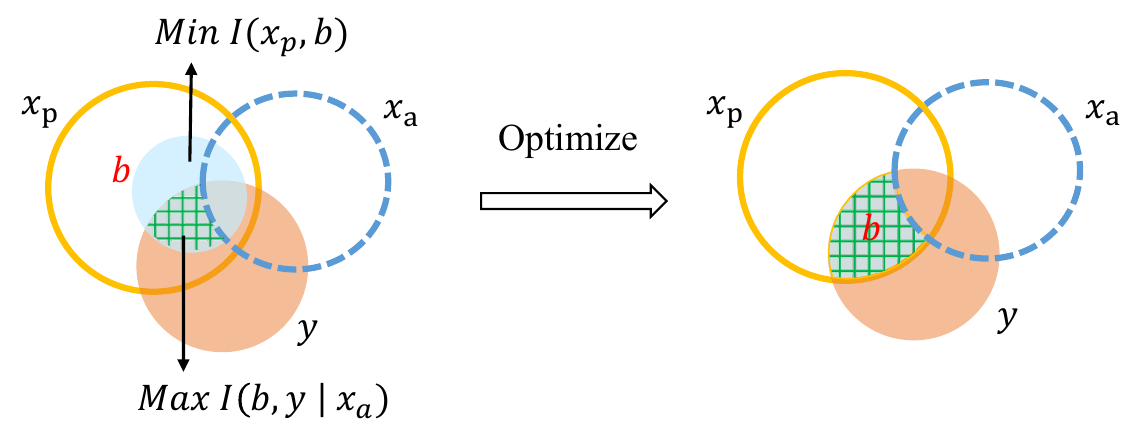}
  \caption{The optimization process of the multimodal conditional information bottleneck is shown. The left is the initial state, where the blue section represents the latent state $b$ generated from the primary modality, and the green cells denote the conditional mutual information between the auxiliary modality, latent state, and target. The right is the ideal state: $b$ contains all information relevant to the target $y$, free from redundancy, and integrates complementary information from the primary modality concerning the auxiliary modality.}
  \label{fig:mm}
\end{figure}

\begin{equation}
    \min_{p(b|x_p, x_a)} \, \underbrace{I(x_p; b)}_{\text{Compress redundancy}}  -  \lambda \, \underbrace{I(b; y \mid x_a)}_{\text{Retain complementarity}},
    \label{eq:2}
\end{equation}
where $I(x_p; b)$ encourages compression of redundancy from $x_p$. $I(b; y \mid x_a)$ ensures $b$ retains complementary information useful for predicting $y$ given $x_a$. $\lambda$ balances the trade-off between compression and retention.

\noindent \textbf{Compress redundancy.} This term penalizes the mutual information between the primary modality $x_p$ and the latent state $b$, encouraging $b$ to be a compressed representation of $x_p$. The mutual information between $x_p$ and $b$ is:
\begin{equation}
    I(x_p; b) = \int p(x_p, b) \log \frac{p(b \mid x_p)}{p(b)} \mathrm{d}x_p \mathrm{d}b.
    \label{eq:3}
\end{equation}
Computing the marginal distribution $p(b)$ is intractable, so we introduce a variational prior $r(b)$ to approximate $p(b)$. Assuming $p(b \mid x_p)$ is approximated by a variational distribution $q(b \mid x_p)$, the upper bound becomes:
\begin{equation}
    I(x_p; b) \leq \mathbb{E}_{p(x_p)} \left[ D_{\text{KL}}(q(b \mid x_p) \| r(b)) \right].
    \label{eq:4}
\end{equation}
We define the loss term for compression as the KL divergence between $q(b \mid x_p)$ and $r(b)$. Specifically, in designing the optimizer: choose a standard Gaussian $\mathcal{N}(0, I)$ as Prior $r(b)$, while utilizing a transformer architecture to capture detailed and relevant information from $x_p$ for simulating $q(b \mid x_p)$. The encoder models $q(b \mid x_p)$ as a Gaussian distribution $\mathcal{N}(\mu(x_p), \sigma^2(x_p))$. The latent variable $b$ is obtained by reparameterizing and sampling from $q(b \mid x_p)$.
The KL divergence is calculated as:
\begin{equation}
    D_{\text{KL}}(q(b \mid x_p) \| r(b)) = \frac{1}{2} \sum_{i=1}^{d} ( \sigma_i^2 + \mu_i^2 - 1 - \log \sigma_i^2 ).
    \label{Eq:5}
\end{equation}
\cref{Eq:5} encourages $b$ to gather the most essential information from $x_p$, effectively compressing the redundancy in $x_p$.

\noindent \textbf{Retain complementarity.} This term encourages $b$ to retain information about $y$ that is complementary to $x_a$. The conditional mutual information is defined as:
\begin{equation}
    I(b; y \mid x_a) = \int \! \! p(x_a, b, y) \log \frac{p(y \mid b, x_a)}{p(y \mid x_a)} \mathrm{d}x_a \mathrm{d}b \mathrm{d}y.
    \label{eq:6}
\end{equation}
Direct computation is intractable. Introducing a variational distribution $q(y \mid b, x_a)$, we obtain:
\begin{equation}
    \begin{aligned}
    I(b;\! y \! \mid \! x_a) &= \mathbb{E}_{p(x_a, b, y)}\left[ \log \frac{p(y\mid b, x_a)}{p(y\mid x_a)} \right] \\
    &\geq \mathbb{E}_{p(x_a, b, y)} [ \log\! q(y\! \mid \! b, x_a) - \log\! p(y\! \mid\! x_a)].
    \label{eq:7}
    \end{aligned}
\end{equation}
Ignoring the constant term $\log p(y \mid x_a)$ (since it does not depend on $b$), we get the lower bound:
\begin{equation}
    I(b; y \mid x_a) \geq \mathbb{E}_{p(x_a, b, y)} \left[ \log q(y \mid b, x_a) \right].
    \label{eq:8}
\end{equation}
We define the loss term for retention as the negative expected log-likelihood. Then the Evidence Lower Bound Objective (ELBO) method is adopted for the variational approximation of $q(y \mid b, x_a)$. Combining $b$ and $x_a$, we construct a transformer-based neural network estimator to model $q(y \mid b, x_a)$. The expected log-likelihood is approximated using samples $b^{(l)}$ drawn from $q(b \mid x_p)$:
\begin{equation}
    \mathbb{E}_{p(x_a, b, y)} \! \left[ \log q(y \mid b, x_a) \right] \approx \frac{1}{L} \! \sum_{l=1}^{L} \log q(y \mid b^{(l)}, x_a),
\end{equation}
where $L$ is the number of samples.

This loss encourages the model to reconstruct $y$ from the combined $b$ and $x_a$, maximizing the conditional mutual information $I(b; y \mid x_a)$. By precisely formulating each step and keeping the derivation concise, we establish a tractable lower bound for $I(b; y \mid x_a)$ using variational inference and ELBO methods, which effectively integrates the auxiliary modality $x_a$ with the latent representation $b$ to enhance predictive performance. The overall loss function is:
\begin{equation}
    \begin{aligned}
        \mathcal{L} = \frac{1}{2} \sum_{i=1}^{d}(\sigma_i^2 + \mu_i^2 - 1 - \log \sigma_i^2 ) \\
        - \lambda \frac{1}{L} \sum_{l=1}^{L}\log q(y \mid b^{(l)}, x_a).
    \end{aligned}
    \label{eq:modified}
\end{equation}

\begin{figure}[t]
  \centering
  \includegraphics[width=0.5\linewidth]{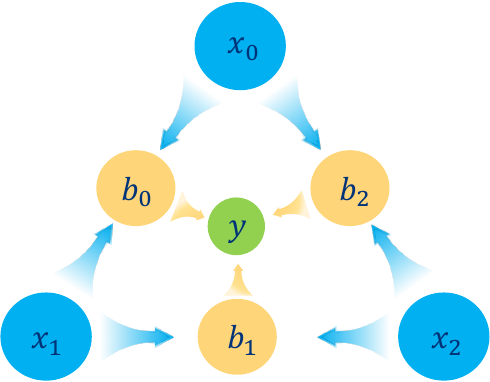}
  \caption{By constructing three latent state $b_0$, $b_1$ and $b_2$, pertinent information transfer between the three modal $x_0$, $x_1$ and $x_2$ is facilitated. Finally, integrated data leads to the prediction of $y$.}
  \label{fig:fig5}
\end{figure}


\subsubsection{Overall Architecture and Optimization}

To uncover latent information within each modality, we design a module for fine-grained feature extraction. Using GENTLE, we align audio and segment audio at the word level, which then serves as a reference for aligning fine-grained features in the visual modality. Next is the multimodal feature fusion module MCIB, which integrates complementary information across modalities while filtering out redundancy. Specifically, the MCIB minimizes the mutual information between the primary modality $x_p$ and the latent state $b$, condensing $x_p$ and filtering out redundant information. Simultaneously, it maximizes the conditional mutual information between the auxiliary modality $x_a$, the latent state $b$, and the target $y$, aiming to incorporate additional relevant information from $x_a$ so that $b$ holds useful information for predicting $y$. The model concatenates the trained latent state $b$ for prediction, which contains the "redundancy-removed, effective complementarity" information distilled through the conditional information bottleneck.

As shown in Figure \ref{fig:fig5}, in the context of multimodal learning, we alternately designate each of the three modalities as the primary and auxiliary modalities to train the fusion framework based on the MCIB algorithm jointly. To better control the degree of information compression, we introduce modality-specific hyperparameters $\lambda_{0}$, $\lambda_{1}$, and $\lambda_{2}$ to balance the conditional information bottleneck loss. Letting 0, 1, and 2 represent the three modalities, their respective loss functions are defined as \cref{e1}:

\begin{equation}
    \begin{aligned}
        \mathcal{L}_0 &= \mathcal{L}_{\text{IB}_0} + \lambda_{0} \mathcal{L}_{\text{CIB}_2}, \\
        \mathcal{L}_1 &= \mathcal{L}_{\text{IB}_1} + \lambda_{1} \mathcal{L}_{\text{CIB}_0}, \\
        \mathcal{L}_2 &= \mathcal{L}_{\text{IB}_2} + \lambda_{2} \mathcal{L}_{\text{CIB}_1}.
    \end{aligned}
    \label{e1}
\end{equation}

Furthermore, to fully exploit the fused latent state $b$ in the conditional information bottleneck, we introduce a prediction loss $\mathcal{L}_{\text{pred}}$ from $b$ to $y$, where \(\alpha_0\), \(\alpha_1\), \(\alpha_2\) and \(\beta\) are the weighting coefficients. The model is trained with the following final objective: $\mathcal{L}_{\text{total}} = \alpha_{0} \mathcal{L}_0 + \alpha_{1} \mathcal{L}_1 + \alpha_{2} \mathcal{L}_2 + \beta \mathcal{L}_{\text{pred}}$.

\section{Experiments}

\textbf{Dataset and Evaluation Metrics.} Experiments were conducted on MUStARD++ and MUStARD++$^{R}$, respectively. Given the balanced class distribution, we report results using weighted precision, weighted recall, and weighted F1-score metrics. In addition, we tested the MCIB method on multimodal sentiment analysis tasks (CMU-MOSI and CMU-MOSEI in the Extended version), achieving highly competitive results, which validate its generalization capability.

\textbf{Implementation Details.} To extract fine-grained features, we utilize several high-performance backbones. Text features are obtained using a pre-trained DeBERTa \cite{he2021debertav3}, yielding a representation size of $d_t = 768$. For audio features, Mel Frequency Cepstral Coefficients (MFCC) and Mel spectrograms are generated with Librosa \cite{McFee2018librosalibrosa0}, along with prosodic features from OpenSMILE 3 \cite{eyben2010opensmile}, resulting in a combined representation of $d_a = 291$. The video features are extracted by processing utterance frames through the pool5 layer of a ResNet-152 \cite{he2016deep} pre-trained on ImageNet \cite{deng2009imagenet}, producing a visual representation of size $d_v = 2048$.  All experiments were conducted on Nvidia A100 GPUs with 40GB of memory. Multiple trials demonstrate that combinations of random hyperparameters within $[1, 64]$ typically result in local optima after 10 searches. The hyperparameter analysis reveals that the most significant interactions occur between $\lambda_{2} \And \alpha_{1}$, $\lambda_1 \And \alpha_{1}$, and $\alpha_{0} \And \alpha_{1}$. The stability analysis highlighted that $\alpha_{1}$ demonstrated the best stability. Experimental results were averaged over five runs.

\subsection{Comparison with baseline models}

\setlength{\tabcolsep}{1pt}
\begin{table}[h]
\centering
\begin{tabular}{lccc}
\hline
Method &Precision & Recall &F1-score \\
\hline
PredGaze$^{\scalebox{1.2}{$\circ$}}$ & 71.06 & 71.06 & 71.06 \\
\hline
VyAnG-Net$^{\scalebox{1.2}{$\circ$}}$ & 72.41 & 72.05 & 72.23 \\
\hline
FIES$^{\scalebox{1.2}{$\circ$}}$ & 71.48 & 71.43 & 71.45 \\
FIES$^{\scalebox{1.2}{$\circ$}\heartsuit}$ & 74.31 & 74.29 & 74.29 \\
\hline
ABCA-IMI$^{\scalebox{1.2}{$\circ$}}$ & 71.90 & 69.90 & 70.90 \\
ABCA-IMI$^{\scalebox{1.2}{$\circ$}\diamondsuit}$ & 76.20 & 74.20 & 75.20 \\
\hline
SIB$^{\scalebox{1.2}{$\circ$}}$ & 69.67 & 69.54 & 69.39 \\
SIB (w/o shortcuts) & 68.30 & 68.26 & 68.21 \\
\hline
DIB$^{\scalebox{1.2}{$\circ$}}$ & 70.96 & 71.12 & 70.98 \\
DIB (w/o shortcuts) & 70.44 & 70.43 & 70.42 \\
\hline
ITHP $^{\scalebox{1.2}{$\circ$}\diamondsuit}$ & 71.39 & 70.95 & 70.93 \\
ITHP (w/o shortcuts) & 68.27 & 68.29 & 68.27 \\
\hline
TBJE$^{\scalebox{1.2}{$\circ$}\heartsuit}$ & 71.79 & 71.90 & 71.82 \\
TBJE (w/o shortcuts) & 69.50 & 68.59 & 68.75 \\
\hline
MUStARD method$^{\scalebox{1.2}{$\circ$}}$ & 71.02 & 70.80 & 70.98 \\
MUStARD method$^{\scalebox{1.2}{$\circ$}\diamondsuit}$ & 71.38 & 71.38 & 71.38 \\
MUStARD method (w/o shortcuts) & 71.07 & 70.03 & 70.03 \\
\hline
SpeechPrompt v2$^{\scalebox{1.2}{$\circ$}}$ & 78.33 & 58.06 & 73.47 \\
SpeechPrompt v2 (w/o shortcuts) & 63.03 & 27.87 & 43.59 \\
\hline
GPT-4o$^{\diamondsuit}$  & 62.66  &  83.90  &  71.74  \\
GPT-4o (w/o shortcuts) & 67.11 & 85.47 & 75.19 \\
\hline
Gemini 2.5$^{\diamondsuit}$ & 62.89  & 84.75 &  72.20 \\
Gemini 2.5 (w/o shortcuts) & 69.12 & 78.99 & 73.73 \\ 
\hline
MCIB$^{\scalebox{1.2}{$\circ$}}$ & 77.18 & 76.30 & 76.85 \\
MCIB$^{\scalebox{1.2}{$\circ$}\diamondsuit}$  & 76.34 & 75.84 & 75.75 \\
MCIB (w/o shortcuts) & 76.14 & 75.83 & 75.64 \\
\hline
\end{tabular}
\caption{
Performance comparison of baseline methods on MUStARD++ and MUStARD++$^{R}$. Superscripts denote the use of specific shortcuts in MUStARD++: \scalebox{1.5}{$\circ$} for canned laughter, $\diamondsuit$ for character, and $\heartsuit$ for emotion. Results labeled "(w/o shortcuts)" correspond to our clean dataset MUStARD++$^{R}$.
}
\label{tab:allMM}
\end{table}

We compare our proposed MCIB with the following baselines on multimodal fusion and sentiment analysis. VyAnG-Net \cite{pandey2025vyang} enhances sarcasm detection by integrating visual-specific attention mechanisms with text captions, while PredGaze \cite{tiwari2023predict} improves sarcasm recognition by utilizing fine-grained visual information such as eye gaze. FIES \cite{gao2024improving} proposes a multimodal approach that integrates audio, textual, sentiment, and emotion data to enhance sarcasm detection, while ABCA-IMI \cite{li2024attention} identifies sarcasm through multiple inconsistency detection mechanisms. Some methods did not release their code; for instance, the VyAnG-Net results were obtained by training on MUStARD and validating on MUStARD++. SIB \cite{mai2022multimodal, chen2023multimodal} using the IB between each single modality and the target, while DIB \cite{zhang2022information} applies the IB to pairs of modalities, enabling back-optimization. ITHP \cite{xiao2024neuro} was designed with a two-layer IB guide to the modality information flow. TBJE \cite{delbrouck2020transformer} is a cross-attention-based model with high generalization for multimodal fusion. MUStARD method \cite{ray2022multimodal} leverages a collaborative gating mechanism for sarcasm and emotion recognition. Moreover, we report baseline results for the pre-trained speech model SpeechPrompt v2 \cite{chang2023speechprompt}, and leading large language models: OpenAI's latest flagship model GPT-4o \cite{openai2024gpt4ocard}, the newest release from Google DeepMind Gemini 2.5 \cite{comanici2025gemini25pushingfrontier}, under multimodal configurations.

As shown in Table \ref{tab:allMM}, our approach achieved the highest F1 scores of $76.85\%$ and $75.64\%$ on the MUStARD++ and MUStARD++$^{R}$ datasets, respectively, outperforming all baseline methods. The results indicate two aspects: the model's reliance on shortcuts (generalization ability) and its capacity to capture truly useful information (effectiveness of multimodal fusion).

First, our method MCIB achieves strong performance without relying on tricks or dataset-specific shortcuts, demonstrating robust generalization across different data conditions. Comparing results on MUStARD++ and MUStARD++$^{R}$, we observe that, after shortcut cues (such as character labels, canned laughter, or emotional inconsistency) are removed, the performance of most conventional methods drops to varying degrees. Interestingly, GPT-4o and Gemini 2.5 perform better in the absence of these cues, possibly because character information introduces noise that misleads LLMs. These findings indicate that many methods are sensitive to shortcuts, while MCIB remains robust.

Second, using MUStARD++$^{R}$, models cannot rely on shortcuts and must depend on architecture design and multimodal fusion strategies. The results demonstrate that MCIB neither relies on nor overfits to shortcut cues and achieves the most effective fusion strategy. In real-world scenarios with limited auxiliary information, our approach outperforms other multimodal fusion methods by more effectively integrating information across modalities.

Additionally, we conducted cross-training experiments between MUStARD++ and MUStARD++$^{R}$ to further investigate the impact of shortcut learning on multimodal sarcasm detection (see Extended version).

\subsection{Modality and Module Ablation Studies}
This section examines modality fusion in MCIB, the impact of the transformer and fine-grained modules, and results from modality ablation and different modality combinations.

\begin{table}[h]
\centering
\begin{tabular}{lccc}
\hline
Method & Precision & Recall & F1-score \\
\hline
w/o Transformer            & 75.02 & 74.58 & 74.32 \\
w/o Fine-Gained            & 71.65 & 71.21 & 71.19 \\
$x_{v}$                    & 69.98 & 70.00 & 69.99 \\
$x_{a}$                    & 70.43 & 69.16 & 68.97 \\
$x_{t}$                    & 70.60 & 70.83 & 70.98 \\
$x_{va}$                   & 72.06 & 72.08 & 72.04 \\
$x_{at}$                   & 73.75 & 73.75 & 73.69 \\
$x_{tv}$                   & 74.92 & 74.17 & 73.77 \\
$x_{vt}+x_{av}+x_{ta}$     & 75.42 & 74.17 & 74.03 \\
$x_{va}+x_{at}+x_{tv}$     & 76.14 & 75.83 & 75.64\\
\hline
\end{tabular}
\caption{Ablation results: using an MLP (w/o Transformer), coarse-grained features (w/o Fine-Grained), modality removal, and varying modality pairs.}
\label{tab:ab}
\end{table}

\noindent \textbf{Transformer architecture.} Our transformer-based encoder captures richer patterns than the MLP, yielding a $1.32\%$ performance gain despite increased computational cost.

\noindent \textbf{Fine-gained module ablation.} We compare fine-grained and coarse-grained feature extraction to investigate their impact on multimodal tasks. The results show that aligning and representing the three modalities at the word level with fine-grained features enhances sarcasm detection.

\noindent \textbf{Modal ablation.} We conduct experiments with all three modalities. Among single modalities, text performs best, while vision performs worst. For dual modalities, performance is highest when text is primary and vision is auxiliary, and lowest when video is primary and audio is auxiliary.

\noindent \textbf{Varying primary and auxiliary modality pairs.} The three modalities yield six possible sequential pairs. To maximize complementary information, we prioritize combinations that include all modalities. This results in two configurations: ${va + at + tv}$ and ${av + vt + ta}$ (where the first modality is primary, and the second is auxiliary). Experimental results indicate that sarcasm detection performs best when visual assists the audio modality, audio assists text, and text assists the visual modality.

\subsection{Visualizations}
\begin{figure}[h!]
  \centering
  \includegraphics[width=0.9\linewidth]{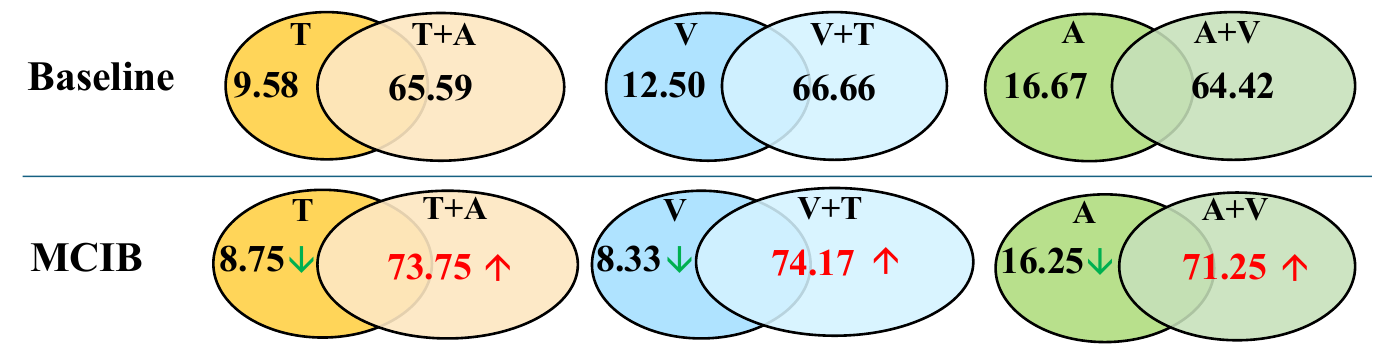}
  \caption{The diagrams illustrate the loss-benefit comparison for adding new modalities to visual, text, and audio modalities. In each Venn diagram, the numbers in each section represent the proportion of correctly predicted samples relative to the total test set.}
  \label{fig:xr}
\end{figure}

Figure \ref{fig:xr} provides an intuitive comparison of the effectiveness of the proposed MCIB method in modality fusion relative to baseline models. The directional trends of the arrows indicate that MCIB optimizes in the correct direction for modality fusion. Observing the overall improvement: misclassifications due to redundant or irrelevant information are significantly reduced across all modality combinations, while complementary information between modalities is effectively utilized. The accuracy of joint predictions from added modalities has also improved.

\section{Conclusion}

This paper first analyzes the shortcut learning problem existing in current multimodal sarcasm detection methods and restructures the task dataset to prevent the learning of shortcut features. Subsequently, we propose a multimodal conditional information bottleneck (MCIB) framework that effectively captures complementary inter-modal information while filtering out irrelevant and misleading redundancies. Extensive experimental results demonstrate that our model uses multimodal data more effectively, achieving state-of-the-art performance on the multimodal sarcasm detection task. In the future, we aim to refine MCIB into an easily integrable plug-in for various backbone models in multimodal sentiment analysis, thereby boosting their performance.

\section{Acknowledgments}
This work was supported by the National Natural Science Foundation of China (No. 62276026).

\bibliography{aaai2026}

@misc{hu2025emobenchmbench,
      title={EmoBench-M: Benchmarking Emotional Intelligence for Multimodal Large Language Models}, 
      author={He Hu and Yucheng Zhou and Lianzhong You and Hongbo Xu and Qianning Wang and Zheng Lian and Fei Richard Yu and Fei Ma and Laizhong Cui},
      year={2025},
      eprint={2502.04424},
      archivePrefix={arXiv},
      primaryClass={cs.CL},
      url={https://arxiv.org/abs/2502.04424}, 
}

@inproceedings{ray2022multimodal,
  title={A Multimodal Corpus for Emotion Recognition in Sarcasm},
  author={Ray, Anupama and Mishra, Shubham and Nunna, Apoorva and Bhattacharyya, Pushpak},
  booktitle={Proceedings of the Thirteenth Language Resources and Evaluation Conference},
  pages={6992--7003},
  year={2022}
}

@article{wang2024cross,
  title={Cross-modal incongruity aligning and collaborating for multi-modal sarcasm detection},
  author={Wang, Jie and Yang, Yan and Jiang, Yongquan and Ma, Minbo and Xie, Zhuyang and Li, Tianrui},
  journal={Information Fusion},
  volume={103},
  pages={102132},
  year={2024},
  publisher={Elsevier}
}

@article{lu2024fact,
  title={Fact-sentiment incongruity combination network for multimodal sarcasm detection},
  author={Lu, Qiang and Long, Yunfei and Sun, Xia and Feng, Jun and Zhang, Hao},
  journal={Information Fusion},
  volume={104},
  pages={102203},
  year={2024},
  publisher={Elsevier}
}

@inproceedings{tiwari2023predict,
  title={Predict and Use: Harnessing Predicted Gaze to Improve Multimodal Sarcasm Detection},
  author={Tiwari, Divyank and Kanojia, Diptesh and Ray, Anupama and Nunna, Apoorva and Bhattacharyya, Pushpak},
  booktitle={Proceedings of the 2023 Conference on Empirical Methods in Natural Language Processing},
  pages={15933--15948},
  year={2023}
}

@article{yue2023knowlenet,
  title={KnowleNet: Knowledge fusion network for multimodal sarcasm detection},
  author={Yue, Tan and Mao, Rui and Wang, Heng and Hu, Zonghai and Cambria, Erik},
  journal={Information Fusion},
  volume={100},
  pages={101921},
  year={2023},
  publisher={Elsevier}
}

@article{zhang2023learning,
  title={Learning multi-task commonness and uniqueness for multi-modal sarcasm detection and sentiment analysis in conversation},
  author={Zhang, Yazhou and Yu, Yang and Zhao, Dongming and Li, Zuhe and Wang, Bo and Hou, Yuexian and Tiwari, Prayag and Qin, Jing},
  journal={IEEE Transactions on Artificial Intelligence},
  year={2023},
  publisher={IEEE}
}

@inproceedings{gao2024improving,
  title={Improving sarcasm detection from speech and text through attention-based fusion exploiting the interplay of emotions and sentiments},
  author={Gao, Xiyuan and Nayak, Shekhar and Coler, Matt},
  booktitle={Proceedings of Meetings on Acoustics},
  volume={54},
  number={1},
  pages={060002},
  year={2024},
  organization={Acoustical Society of America}
}

@article{li2024attention,
  title={An attention-based, context-aware multimodal fusion method for sarcasm detection using inter-modality inconsistency},
  author={Li, Yangyang and Li, Yuelin and Zhang, Shihuai and Liu, Guangyuan and Chen, Yanqiao and Shang, Ronghua and Jiao, Licheng},
  journal={Knowledge-Based Systems},
  volume={287},
  pages={111457},
  year={2024},
  publisher={Elsevier}
}

@article{pandey2025vyang,
  title={VyAnG-Net: A novel multi-modal sarcasm recognition model by uncovering visual, acoustic and glossary features},
  author={Pandey, Ananya and Vishwakarma, Dinesh Kumar},
  journal={Intelligent Data Analysis},
  pages={1088467X251315637},
  year={2025},
  publisher={SAGE Publications Sage UK: London, England}
}

@inproceedings{chauhan2020sentiment,
  title={Sentiment and emotion help sarcasm? A multi-task learning framework for multi-modal sarcasm, sentiment and emotion analysis},
  author={Chauhan, Dushyant Singh and Dhanush, SR and Ekbal, Asif and Bhattacharyya, Pushpak},
  booktitle={Proceedings of the 58th annual meeting of the association for computational linguistics},
  pages={4351--4360},
  year={2020}
}

@inproceedings{song2024utterance,
  title={Utterance-Level Incongruity Learning Network for Multimodal Sarcasm Detection},
  author={Song, Liujing and Zhao, Zefang and Ma, Yuxiang and Liu, Yuyang and Li, Jun},
  booktitle={2024 26th International Conference on Advanced Communications Technology (ICACT)},
  pages={43--49},
  year={2024},
  organization={IEEE}
}

@inproceedings{liang2022multi,
  title={Multi-modal sarcasm detection via cross-modal graph convolutional network},
  author={Liang, Bin and Lou, Chenwei and Li, Xiang and Yang, Min and Gui, Lin and He, Yulan and Pei, Wenjie and Xu, Ruifeng},
  booktitle={Proceedings of the 60th Annual Meeting of the Association for Computational Linguistics (Volume 1: Long Papers)},
  volume={1},
  pages={1767--1777},
  year={2022},
  organization={Association for Computational Linguistics}
}

@inproceedings{wen2023dip,
  title={Dip: Dual incongruity perceiving network for sarcasm detection},
  author={Wen, Changsong and Jia, Guoli and Yang, Jufeng},
  booktitle={Proceedings of the IEEE/CVF Conference on Computer Vision and Pattern Recognition},
  pages={2540--2550},
  year={2023}
}

@article{ding2023robust,
  title={Robust Multi-Agent Communication With Graph Information Bottleneck Optimization},
  author={Ding, Shifei and Du, Wei and Ding, Ling and Zhang, Jian and Guo, Lili and An, Bo},
  journal={IEEE Transactions on Pattern Analysis and Machine Intelligence},
  year={2023},
  publisher={IEEE}
}

@inproceedings{castro2019towards,
  title={Towards Multimodal Sarcasm Detection (An \_Obviously\_ Perfect Paper)},
  author={Castro, Santiago and Hazarika, Devamanyu and P{\'e}rez-Rosas, Ver{\'o}nica and Zimmermann, Roger and Mihalcea, Rada and Poria, Soujanya},
  booktitle={Proceedings of the 57th Annual Meeting of the Association for Computational Linguistics},
  pages={4619--4629},
  year={2019}
}

@inproceedings{tomar2023your,
  title={Your tone speaks louder than your face! Modality Order Infused Multi-modal Sarcasm Detection},
  author={Tomar, Mohit and Tiwari, Abhisek and Saha, Tulika and Saha, Sriparna},
  booktitle={Proceedings of the 31st ACM International Conference on Multimedia},
  pages={3926--3933},
  year={2023}
}

@inproceedings{tomar2024action,
  title={Action and Reaction Go Hand in Hand! a Multi-modal Dialogue Act Aided Sarcasm Identification},
  author={Tomar, Mohit Singh and Saha, Tulika and Tiwari, Abhisek and Saha, Sriparna},
  booktitle={Proceedings of the 2024 Joint International Conference on Computational Linguistics, Language Resources and Evaluation (LREC-COLING 2024)},
  pages={298--309},
  year={2024}
}

@article{zhang2024self,
  title={Self-Adaptive Representation Learning Model for Multi-Modal Sentiment and Sarcasm Joint Analysis},
  author={Zhang, Yazhou and Yu, Yang and Wang, Mengyao and Huang, Min and Hossain, M Shamim},
  journal={ACM Transactions on Multimedia Computing, Communications and Applications},
  volume={20},
  number={5},
  pages={1--17},
  year={2024},
  publisher={ACM New York, NY}
}

@article{liu2023quantum,
  title={A quantum probability driven framework for joint multi-modal sarcasm, sentiment and emotion analysis},
  author={Liu, Yaochen and Zhang, Yazhou and Song, Dawei},
  journal={IEEE Transactions on Affective Computing},
  volume={15},
  number={1},
  pages={326--341},
  year={2023},
  publisher={IEEE}
}

@article{tiwari2024quantum,
  title={Quantum fuzzy neural network for multimodal sentiment and sarcasm detection},
  author={Tiwari, Prayag and Zhang, Lailei and Qu, Zhiguo and Muhammad, Ghulam},
  journal={Information Fusion},
  volume={103},
  pages={102085},
  year={2024},
  publisher={Elsevier}
}

@article{helal2024contextual,
  title={A contextual-based approach for sarcasm detection},
  author={Helal, Nivin A and Hassan, Ahmed and Badr, Nagwa L and Afify, Yasmine M},
  journal={Scientific Reports},
  volume={14},
  number={1},
  pages={15415},
  year={2024},
  publisher={Nature Publishing Group UK London}
}

@misc{xiao2024neuro,
      title={Neuro-Inspired Information-Theoretic Hierarchical Perception for Multimodal Learning}, 
      author={Xiongye Xiao and Gengshuo Liu and Gaurav Gupta and Defu Cao and Shixuan Li and Yaxing Li and Tianqing Fang and Mingxi Cheng and Paul Bogdan},
      year={2024},
      eprint={2404.09403},
      archivePrefix={arXiv},
      primaryClass={cs.LG},
      url={https://arxiv.org/abs/2404.09403}, 
}

@inproceedings{zhang2022information,
  title={Information Bottleneck based Representation Learning for Multimodal Sentiment Analysis},
  author={Zhang, Tonghui and Zhang, Haiying and Xiang, Shuke and Wu, Tong},
  booktitle={Proceedings of the 6th International Conference on Control Engineering and Artificial Intelligence},
  pages={7--11},
  year={2022}
}

@article{mai2022multimodal,
  title={Multimodal information bottleneck: Learning minimal sufficient unimodal and multimodal representations},
  author={Mai, Sijie and Zeng, Ying and Hu, Haifeng},
  journal={IEEE Transactions on Multimedia},
  volume={25},
  pages={4121--4134},
  year={2022},
  publisher={IEEE}
}

@inproceedings{chen2023multimodal,
  title={Multimodal Sentiment Analysis Based on Information Bottleneck and Attention Mechanisms},
  author={Chen, Xiangrui and Wu, Zhendong and Tang, Yu and Han, Rong},
  booktitle={2023 2nd International Conference on Cloud Computing, Big Data Application and Software Engineering (CBASE)},
  pages={150--156},
  year={2023},
  organization={IEEE}
}

@article{liu2024multimodal,
  title={Multimodal consistency-specificity fusion based on information bottleneck for sentiment analysis},
  author={Liu, Wei and Cao, Shenchao and Zhang, Sun},
  journal={Journal of King Saud University-Computer and Information Sciences},
  volume={36},
  number={2},
  pages={101943},
  year={2024},
  publisher={Elsevier}
}

@inproceedings{zhang2022unimodal,
  title={Unimodal and Multimodal Integrated Representation Learning via Improved Information Bottleneck for Multimodal Sentiment Analysis},
  author={Zhang, Tonghui and Dong, Changfei and Su, Jinsong and Zhang, Haiying and Li, Yuzheng},
  booktitle={CCF International Conference on Natural Language Processing and Chinese Computing},
  pages={564--576},
  year={2022},
  organization={Springer}
}

@article{yan2023cross,
  title={Cross-modal clustering with deep correlated information bottleneck method},
  author={Yan, Xiaoqiang and Mao, Yiqiao and Ye, Yangdong and Yu, Hui},
  journal={IEEE Transactions on Neural Networks and Learning Systems},
  year={2023},
  publisher={IEEE}
}

@inproceedings{alemi2022deep,
  title={Deep Variational Information Bottleneck},
  author={Alemi, Alexander A and Fischer, Ian and Dillon, Joshua V and Murphy, Kevin},
  booktitle={International Conference on Learning Representations},
  year={2022}
}

@article{slonim1999agglomerative,
  title={Agglomerative information bottleneck},
  author={Slonim, Noam and Tishby, Naftali},
  journal={Advances in neural information processing systems},
  volume={12},
  year={1999}
}

@article{wyner1978definition,
  title={A definition of conditional mutual information for arbitrary ensembles},
  author={Wyner, Aaron D},
  journal={Information and Control},
  volume={38},
  number={1},
  pages={51--59},
  year={1978},
  publisher={Elsevier}
}

@article{fleuret2004fast,
  title={Fast binary feature selection with conditional mutual information.},
  author={Fleuret, Fran{\c{c}}ois},
  journal={Journal of Machine learning research},
  volume={5},
  number={9},
  year={2004}
}

@misc{
li2023modality,
title={Modality Complementariness: Towards Understanding Multi-modal Robustness},
author={Siting Li and Chenzhuang Du and Yu Huang and Longbo Huang and Hang Zhao},
year={2023},
url={https://openreview.net/forum?id=gfHLOC35Zh}
}

@article{he2024efficient,
  title={Efficient Modality Selection in Multimodal Learning},
  author={He, Yifei and Cheng, Runxiang and Balasubramaniam, Gargi and Tsai, Yao-Hung Hubert and Zhao, Han},
  journal={Journal of Machine Learning Research},
  volume={25},
  number={47},
  pages={1--39},
  year={2024}
}

@article{ji2022increasing,
  title={Increasing visual awareness in multimodal neural machine translation from an information theoretic perspective},
  author={Ji, Baijun and Zhang, Tong and Zou, Yicheng and Hu, Bojie and Shen, Si},
  journal={arXiv preprint arXiv:2210.08478},
  year={2022}
}

@inproceedings{gondek2003conditional,
  title={Conditional information bottleneck clustering},
  author={Gondek, David and Hofmann, Thomas},
  booktitle={3rd ieee international conference on data mining, workshop on clustering large data sets},
  pages={36--42},
  year={2003}
}

@inproceedings{zadeh2017tensor,
  title={Tensor Fusion Network for Multimodal Sentiment Analysis},
  author={Zadeh, Amir and Chen, Minghai and Poria, Soujanya and Cambria, Erik and Morency, Louis-Philippe},
  booktitle={Proceedings of the 2017 Conference on Empirical Methods in Natural Language Processing},
  pages={1103--1114},
  year={2017}
}

@article{joshi2017automatic,
  title={Automatic sarcasm detection: A survey},
  author={Joshi, Aditya and Bhattacharyya, Pushpak and Carman, Mark J},
  journal={ACM Computing Surveys (CSUR)},
  volume={50},
  number={5},
  pages={1--22},
  year={2017},
  publisher={ACM New York, NY, USA}
}

@article{molavipour2021neural,
  title={Neural estimators for conditional mutual information using nearest neighbors sampling},
  author={Molavipour, Sina and Bassi, Germ{\'a}n and Skoglund, Mikael},
  journal={IEEE transactions on signal processing},
  volume={69},
  pages={766--780},
  year={2021},
  publisher={IEEE}
}

@article{geirhos2020shortcut,
  title={Shortcut learning in deep neural networks},
  author={Geirhos, Robert and Jacobsen, J{\"o}rn-Henrik and Michaelis, Claudio and Zemel, Richard and Brendel, Wieland and Bethge, Matthias and Wichmann, Felix A},
  journal={Nature Machine Intelligence},
  volume={2},
  number={11},
  pages={665--673},
  year={2020},
  publisher={Nature Publishing Group UK London}
}

@inproceedings{McFee2018librosalibrosa0,
  title={librosa/librosa: 0.6.0},
  author={Brian McFee and Matt McVicar and Stefan Balke and Carl Thom{\'e} and Colin Raffel and Oriol Nieto and Eric Battenberg and Daniel P. W. Ellis and Ryuichi Yamamoto and Josh Moore and Rachel M. Bittner and Keunwoo Choi and Fabian-Robert St{\"o}ter and Siddhartha Kumar and Simon Waloschek and Seth and Rimvydas Naktinis and Douglas Repetto and Curtis Fjord Hawthorne and Cj Carr and hojinlee and Waldir Pimenta and Petr Viktorin and Paul Brossier and Jo{\~a}o Felipe Santos and JackieWu and Erik and Adrian Holovaty},
  year={2018},
  url={https://api.semanticscholar.org/CorpusID:188532486}
}

@article{he2021debertav3,
  title={Debertav3: Improving deberta using electra-style pre-training with gradient-disentangled embedding sharing},
  author={He, Pengcheng and Gao, Jianfeng and Chen, Weizhu},
  journal={arXiv preprint arXiv:2111.09543},
  year={2021}
}

@inproceedings{eyben2010opensmile,
  title={Opensmile: the munich versatile and fast open-source audio feature extractor},
  author={Eyben, Florian and W{\"o}llmer, Martin and Schuller, Bj{\"o}rn},
  booktitle={Proceedings of the 18th ACM international conference on Multimedia},
  pages={1459--1462},
  year={2010}
}

@inproceedings{he2016deep,
  title={Deep residual learning for image recognition},
  author={He, Kaiming and Zhang, Xiangyu and Ren, Shaoqing and Sun, Jian},
  booktitle={Proceedings of the IEEE conference on computer vision and pattern recognition},
  pages={770--778},
  year={2016}
}

@inproceedings{deng2009imagenet,
  title={Imagenet: A large-scale hierarchical image database},
  author={Deng, Jia and Dong, Wei and Socher, Richard and Li, Li-Jia and Li, Kai and Fei-Fei, Li},
  booktitle={2009 IEEE conference on computer vision and pattern recognition},
  pages={248--255},
  year={2009},
  organization={Ieee}
}

@article{delbrouck2020transformer,
  title={A Transformer-based joint-encoding for Emotion Recognition and Sentiment Analysis},
  author={Delbrouck, Jean-Benoit and Tits, No{\'e} and Brousmiche, Mathilde and Dupont, St{\'e}phane},
  journal={ACL 2020},
  pages={1},
  year={2020}
}

@article{chang2023speechprompt,
  title={Speechprompt v2: Prompt tuning for speech classification tasks},
  author={Chang, Kai-Wei and Wang, Yu-Kai and Shen, Hua and Kang, Iu-thing and Tseng, Wei-Cheng and Li, Shang-Wen and Lee, Hung-yi},
  journal={arXiv preprint arXiv:2303.00733},
  year={2023}
}

@article{arora2023universlu,
  title={Universlu: Universal spoken language understanding for diverse classification and sequence generation tasks with a single network},
  author={Arora, Siddhant and Futami, Hayato and Jung, Jee-weon and Peng, Yifan and Sharma, Roshan and Kashiwagi, Yosuke and Tsunoo, Emiru and Watanabe, Shinji},
  journal={arXiv preprint arXiv:2310.02973},
  year={2023}
}

@inproceedings{shah2022emotion,
  title={Emotion enriched retrofitted word embeddings},
  author={Shah, Sapan and Reddy, Sreedhar and Bhattacharyya, Pushpak},
  booktitle={Proceedings of the 29th International Conference on Computational Linguistics},
  pages={4136--4148},
  year={2022}
}

@misc{openai2024gpt4ocard,
      title={GPT-4o System Card}, 
      author={OpenAI et al.},
      year={2024},
      eprint={2410.21276},
      archivePrefix={arXiv},
      primaryClass={cs.CL},
      url={https://arxiv.org/abs/2410.21276}, 
}

@misc{comanici2025gemini25pushingfrontier,
      title={Gemini 2.5: Pushing the Frontier with Advanced Reasoning, Multimodality, Long Context, and Next Generation Agentic Capabilities}, 
      author={Google DeepMind et al.},
      year={2025},
      eprint={2507.06261},
      archivePrefix={arXiv},
      primaryClass={cs.CL},
      url={https://arxiv.org/abs/2507.06261}, 
}

@article{li2023mia,
  title={MIA-Net: Multi-modal interactive attention network for multi-modal affective analysis},
  author={Li, Shuzhen and Zhang, Tong and Chen, Bianna and Chen, CL Philip},
  journal={IEEE Transactions on Affective Computing},
  volume={14},
  number={4},
  pages={2796--2809},
  year={2023},
  publisher={IEEE}
}

@article{yue2024mutual,
  title={Mutual Information of Crossmodal Utterance Representation for Multimodal Sentiment Analysis},
  author={Yue, Dong and Wei, Xiangsen and others},
  journal={IEEE Transactions on Affective Computing},
  year={2024},
  publisher={IEEE}
}

@inproceedings{yang2023semantic,
  title={Semantic Interaction Fusion Framework for Multimodal Sentiment Recognition},
  author={Yang, Shanliang and Cui, Lichao and Wang, Tao},
  booktitle={2023 IEEE International Conference on Systems, Man, and Cybernetics (SMC)},
  pages={2132--2137},
  year={2023},
  organization={IEEE}
}

@inproceedings{zadeh2018multimodal,
  title={Multimodal language analysis in the wild: Cmu-mosei dataset and interpretable dynamic fusion graph},
  author={Zadeh, AmirAli Bagher and Liang, Paul Pu and Poria, Soujanya and Cambria, Erik and Morency, Louis-Philippe},
  booktitle={Proceedings of the 56th Annual Meeting of the Association for Computational Linguistics (Volume 1: Long Papers)},
  pages={2236--2246},
  year={2018}
}

@INPROCEEDINGS{10439124,
  author={Chen, Xiangrui and Wu, Zhendong and Tang, Yu and Han, Rong},
  booktitle={2023 2nd International Conference on Cloud Computing, Big Data Application and Software Engineering (CBASE)}, 
  title={Multimodal Sentiment Analysis Based on Information Bottleneck and Attention Mechanisms}, 
  year={2023},
  volume={},
  number={},
  pages={150-156},
  doi={10.1109/CBASE60015.2023.10439124}}

@article{huang2023dominant,
  title={Dominant SIngle-Modal SUpplementary Fusion (SIMSUF) For Multimodal Sentiment Analysis},
  author={Huang, Jian and Ji, Yanli and Qin, Zhen and Yang, Yang and Shen, Heng Tao},
  journal={IEEE Transactions on Multimedia},
  year={2023},
  publisher={IEEE}
}

@inproceedings{liu2024progressive,
  title={Progressive Fusion Network with Mixture of Experts for Multimodal Sentiment Analysis},
  author={Liu, Dahuang and Yang, Zhenguo and Guo, Zhiwei},
  booktitle={2024 16th International Conference on Advanced Computational Intelligence (ICACI)},
  pages={150--157},
  year={2024},
  organization={IEEE}
}

\section{Appendix}
\appendix

To address the inefficiency of multimodal fusion and the existing shortcut learning within sarcasm detection, we propose a Multimodal fusion algorithm based on the Conditional Information Bottleneck (MCIB). The supplementary material is organized into the following five sections:
\begin{enumerate}[label=\Alph*.]  
    \item Detailed theoretical derivation of the MCIB formulation, including all intermediate steps.
    \item Generalization evaluation of MCIB on additional multimodal sentiment analysis tasks.
    \item Experimental details on model configurations, hyperparameter settings, and evaluation intervals.
    \item Cross-dataset generalization evaluation for shortcut learning reveals the varying sensitivity of models to shortcuts and their core ability to extract information through multimodal fusion.
    \item Rethinking and visual analysis of multimodal fusion effectiveness show that some models fail to leverage fusion, while our MCIB method better removes redundancy and enhances complementarity, as demonstrated by ablation and visual comparisons.
\end{enumerate}

\subsection{A. Theoretical Derivation}
\label{sec: Formula Derivation}
The MCIB effectively compresses redundant information between modalities that may lead to misclassification while focusing on extracting complementary information that is useful for the target across different modalities. 

As shown in \cref{eq1}, the objective is to achieve effective information fusion between modality pairs. Here, $x_p$ represents the primary modality, $x_a$ is the auxiliary modality, $y$ is the prediction target, and $b$ is the intermediate state capturing useful information optimized through the CIB process. The first term in the equation represents the compression of redundancy in $x_p$ to the greatest extent, while the second term ensures the retention of complementary information between $x_p$ and $x_a$ as much as possible.

This section presents the derivation of the core formulas \cref{eq1} underlying the proposed MCIB method.

\begin{equation}
    \min_{p(b|x_p, x_a)} \, \underbrace{I(x_p; b)}_{\text{Compress redundancy}}  -  \lambda \, \underbrace{I(b; y \mid x_a)}_{\text{Retain complementarity}}
    \label{eq1}
\end{equation}

\subsubsection{A.1 Compress Redundancy}
We begin with the definition of mutual information:
\begin{align}
    I(x_p; b) &= \int p(x_p, b) \log \frac{p(x_p, b)}{p(x_p)p(b)} \, dx_p \, db \nonumber \\
              &= \int p(x_p, b) \log \frac{p(b \mid x_p)}{p(b)} \, dx_p \, db \nonumber \\
              &= \int p(x_p) p(b \mid x_p) \log \frac{p(b \mid x_p)}{p(b)} \, dx_p \, db.
    \label{eq:2}
\end{align}
Since directly calculating $p(b)$ is challenging, we introduce a variational prior distribution $r(b)$ to approximate $p(b)$. Using the properties of logarithms, the logarithmic term can be split as:
\begin{equation}
    \log \frac{p(b \mid x_p)}{p(b)} = \log \frac{p(b \mid x_p)}{r(b)} - \log \frac{p(b)}{r(b)}.
    \label{eq3}
\end{equation}
Substituting this into the mutual information, we obtain:
\begin{equation}
    \begin{split}
        I(x_p; b) &= \int p(x_p) p(b \mid x_p) \bigg[ \log \frac{p(b \mid x_p)}{r(b)} \\
        &\quad - \log \frac{p(b)}{r(b)} \bigg] dx_p \, db \\
        &= \int p(x_p) p(b \mid x_p) \log \frac{p(b \mid x_p)}{r(b)} \, dx_p \, db \\
        &\quad - \int p(x_p) p(b \mid x_p) \log \frac{p(b)}{r(b)} \, dx_p \, db.
    \end{split}
    \label{eq:split}
\end{equation}
Since $\int p(x_p) p(b \mid x_p) \, dx_p = p(b)$:
\begin{equation}
    \begin{split}
        \int p(x_p) p(b \mid x_p) \log \frac{p(b)}{r(b)} \, dx_p \, db 
        &= \int p(b) \log \frac{p(b)}{r(b)} \, db \\
        &= D_{\text{KL}}(p(b) \| r(b)).
    \end{split}
\end{equation}
Thus, we derive \cref{eq3}:
\begin{equation}
    \begin{split}
        I(x_p; b) = & \int p(x_p) p(b \mid x_p) \log \frac{p(b \mid x_p)}{r(b)} \, dx_p \, db \\
        & - D_{\text{KL}}(p(b) \| r(b)).
    \end{split}
\end{equation}
Since KL divergence is always non-negative, $D_{\text{KL}}(p(b) \| r(b)) \geq 0$:
\begin{equation}
    I(x_p; b) \leq \int p(x_p) p(b \mid x_p) \log \frac{p(b \mid x_p)}{r(b)} \, dx_p \, db.
\end{equation}
For a fixed $x_p$, the KL divergence between $p(b \mid x_p)$ and $r(b)$ is given by:
\begin{equation}
    D_{\text{KL}}(p(b \mid x_p) \| r(b)) = \int p(b \mid x_p) \log \frac{p(b \mid x_p)}{r(b)} \, db.
\end{equation}
Therefore, the upper bound of mutual information can be expressed as:
\begin{equation}
    I(x_p; b) \leq \int p(x_p) D_{\text{KL}}(p(b \mid x_p) \| r(b)) \, dx_p.
\end{equation}
To simplify the computation of $p(b \mid x_p)$, we approximate it with an encoder $q(b \mid x_p)$. The mutual information upper bound then becomes:
\begin{equation}
    \begin{split}
        I(x_p; b) &\leq \int p(x_p) D_{\text{KL}}(q(b \mid x_p) \| r(b)) \, dx_p \\
        &= \mathbb{E}_{p(x_p)} \left[ D_{\text{KL}}(q(b \mid x_p) \| r(b)) \right]
    \end{split}
\end{equation}
We define the mutual information upper bound as the compression loss term:
\begin{equation}
    \mathcal{L}_{\text{IB}} = \mathbb{E}_{p(x_p)} \left[ D_{\text{KL}}(q(b \mid x_p) \| r(b)) \right].
    \label{eq5}
\end{equation}
This loss penalizes the mutual information between $x_p$ and $b$, encouraging $b$ to serve as a compressed representation of $x_p$.

\subsubsection{A.2 Retain Complementarity}
The conditional mutual information is:
\begin{equation}
    \begin{split}
        I(b; y \mid x_a) 
        &= \int p(x_a, b, y) \log \frac{p(b, y \mid x_a)}{p(b \mid x_a)p(y \mid x_a)} \, \mathrm{d}x_a \, \mathrm{d}b \, \mathrm{d}y \\
        &= \int p(x_a, b, y) \log \frac{p(b \mid x_a)p(y \mid b, x_a)}{p(b \mid x_a)p(y \mid x_a)} \, \mathrm{d}x_a \, \mathrm{d}b \, \mathrm{d}y \\
        &= \int p(x_a, b, y) \log \frac{p(y \mid b, x_a)}{p(y \mid x_a)} \, \mathrm{d}x_a \, \mathrm{d}b \, \mathrm{d}y
    \end{split}
\end{equation}
The conditional mutual information can be rewritten in an expectation form:
\begin{equation}
    I(b; y \mid x_a) = \mathbb{E}_{p(x_a, b, y)}\left[ \log \frac{p(y \mid b, x_a)}{p(y \mid x_a)} \right].
\end{equation}
Since directly computing $p(y \mid b, x_a)$ and $p(y \mid x_a)$ can be intractable, we introduce a variational distribution $q(y \mid b, x_a)$ to approximate $p(y \mid b, x_a)$. Using the log inequality and the non-negativity of KL divergence:
\begin{align}
    & D_{\text{KL}}(p(y \mid b, x_a) \| q(y \mid b, x_a)) \nonumber \\
    &= \mathbb{E}_{p(y \mid b, x_a)}\left[ \log \frac{p(y \mid b, x_a)}{q(y \mid b, x_a)} \right] \geq 0.
\end{align}
Thus:
\begin{equation}
    \mathbb{E}_{p(y \mid b, x_a)}\left[ \log p(y \mid b, x_a) \right] \geq \mathbb{E}_{p(y \mid b, x_a)}\left[ \log q(y \mid b, x_a) \right].
\end{equation}
Substituting this into the expression for conditional mutual information, we have:
\begin{equation}
    \begin{aligned}
        I(b; y \mid x_a) &= \mathbb{E}_{p(x_a, b)}\left[ \mathbb{E}_{p(y \mid b, x_a)}\left[ \log \frac{p(y \mid b, x_a)}{p(y \mid x_a)} \right] \right] \\
        &= \mathbb{E}_{p(x_a, b, y)}\left[ \log \frac{p(y \mid b, x_a)}{p(y \mid x_a)} \right] \\
        &\geq \mathbb{E}_{p(x_a, b, y)}\left[ \log q(y \mid b, x_a) - \log p(y \mid x_a) \right].
    \end{aligned}
\end{equation}
Note that $\log p(y \mid x_a)$ is independent of $b$ and can be treated as a constant during optimization. To simplify, we can ignore this term:
\begin{equation}
    I(b; y \mid x_a) \geq \mathbb{E}_{p(x_a, b, y)}\left[ \log q(y \mid b, x_a) \right].
    \label{eq:11}
\end{equation}
The expectation can be approximated as:
\begin{equation}
    \mathbb{E}_{q(b \mid x_p)}\left[ \mathbb{E}_{p(x_a, y)}\left[ \log q(y \mid b, x_a) \right] \right].
\end{equation}
Using a finite number of samples $L$, we approximate the expectation as:
\begin{equation}
    \mathbb{E}_{q(b \mid x_p)}\left[ \log q(y \mid b, x_a) \right] \approx \frac{1}{L} \sum_{l=1}^{L} \log q(y \mid b^{(l)}, x_a),
    \label{eq:14}
\end{equation}
where $b^{(l)}$ is sampled from $q(b \mid x_p)$.
The loss term for the Conditional Information Bottleneck (CIB) can be defined as:
\begin{equation}
   \mathcal{L}_{\text{CIB}} = -\mathbb{E}_{p(x_a, b, y)}\left[ \log q(y \mid b, x_a) \right].
\end{equation}
To compute the expectations and enable backpropagation through stochastic variables, we use the reparameterization trick:
\begin{equation}
   b = \mu(x_p) + \sigma(x_p) \odot \epsilon, \quad \epsilon \sim \mathcal{N}(0, I).
\end{equation}
The expectation is approximated using Monte Carlo samples:
\begin{equation}
   \mathbb{E}_{q(b \mid x_p)}\left[ \log q(y \mid b, x_a) \right] \approx \frac{1}{L}\sum_{l=1}^{L} \log q(y \mid b^{(l)}, x_a),
\end{equation}
where $b^{(l)}$ are samples generated using the reparameterization.
Estimating $q(y \mid b, x_a)$: The variational distribution $q(y \mid b, x_a)$ is modeled using a neural network (e.g., a softmax classifier). The objective combines reconstruction loss and KL divergence:
\begin{equation}
   \mathcal{L} = \mathcal{L}_{\text{recon}} + D_{\text{KL}}(q(y \mid b, x_a) \parallel q(y)).
\end{equation}
This corresponds to the ELBO, where:
\begin{itemize}
   \item $\mathcal{L}_{\text{recon}}$ measures how well $q(y \mid b, x_a)$ can reconstruct $y$.
   \item $D_{\text{KL}}$ regularizes $q(y \mid b, x_a)$ to remain close to the prior $q(y)$.
\end{itemize}

\subsection{B. Generalization Evaluation}
\label{sec: Generalization MSA}
To evaluate the generalization ability of our proposed method, we conducted tests on multimodal sentiment recognition using the two famous representative datasets CMU-MOSI and CMU-MOSEI \cite{zadeh2018multimodal}. The evaluation metrics include Mean Absolute Error (MAE), Correlation Coefficient (Corr), Accuracy (Acc), and F1-score (F1).

We selected baselines for comparison based on the following criteria: (1) published within the last two years to ensure the methods are up-to-date, and (2) sourced from prestigious journals or conferences to guarantee the methods' quality and competitiveness.

\begin{table}[h]
  \setlength{\tabcolsep}{3pt} 
  \small
  \centering
  \begin{tabularx}{\linewidth}{Xcccc}
    \toprule
    Methods & MAE$\downarrow$ & Corr$\uparrow$ & Acc$\uparrow$ & F1 $\uparrow$ \\
    \midrule
    PFME~\cite{liu2024progressive}  & 0.873 & 0.713 & 83.38 & 83.38 \\
    SIMSUF~\cite{huang2023dominant} & 0.709 & 0.802 & 86.08 & 85.98 \\
    IBCAM~\cite{10439124}           & 0.713 & 0.796 & 85.80 & 85.80 \\
    SIFF~\cite{yang2023semantic}    & 0.689 & 0.813 & 87.20 & 87.80 \\
    MIUR~\cite{yue2024mutual}       & 0.725 & 0.795 & 85.10 & 85.10 \\
    ITHP~\cite{xiao2024neuro}       & 0.643 & 0.852 & 88.70 & 88.60 \\
    MIA-NET~\cite{li2023mia}        & 0.55  & -     & 89.02 & 88.97 \\
    MCIB (Ours)                     & 0.682 & 0.845 & 89.01 & 89.00 \\
    \bottomrule
  \end{tabularx}
  \caption{Comparison of performance on the CMU-MOSI dataset. $\uparrow$/$\downarrow$ denotes the direction of improvement.}
  \label{tab:mosi_transposed}
\end{table}

\begin{table}[h]
  \centering
  \setlength{\tabcolsep}{3pt} 
  \small
  \begin{tabularx}{\linewidth}{Xcccc}
    \toprule
    Methods & MAE$\downarrow$ & Corr$\uparrow$ & Acc$\uparrow$ & F1 $\uparrow$ \\
    \midrule
    PFME \cite{liu2024progressive} &0.549 & 0.750 & 85.60 & 85.46\\
    SIMSUF\cite{huang2023dominant}  & 0.529 & 0.772 & 86.23 & 86.12 \\
    IBCAM \cite{10439124}  & 0.598 & 0.788 & 86.60 & 86.50 \\
    SIFF \cite{yang2023semantic}  & 0.563 & 0.792 & 86.50 & 86.20 \\
    MIUR \cite{yue2024mutual}  & 0.533 & 0.762 & 85.40 & 85.40 \\
    ITHP \cite{xiao2024neuro}  & 0.564 & 0.813 & 87.30 & 87.40 \\
    MIA-NET \cite{li2023mia}         & 0.49 & - & 88.59 & 88.56 \\
    MCIB (Ours)  & 0.569 & 0.809  &  86.77 &  86.76 \\
    \bottomrule
  \end{tabularx}
  \caption{Compare performance for CMU-MOSEI datasets.}
  \label{tab:mosei_transposed}
\end{table}

PFME incorporates a Progressive Attention Fusion (PAF) module to iteratively extract modal-shared representations using a shared query, and a Mixtures of Attention Experts (MAE) module to generate modality-specific representations through exclusive routing. Three loss functions—contrastive loss for inter-modal associations, orthogonal loss to maintain query invariance, and balance loss for expert assignment—further improve performance.

SIMSUF enhances multimodal sentiment analysis by determining a dominant modality to supplement others, enabling representative feature learning. It further employs a two-branch enhancement module to capture common representations and modality-specific differences, followed by a fusion module to integrate features effectively.

IBCAM addresses the challenges of noise, redundancy, and limited intermodal interaction in existing multimodal sentiment analysis approaches. Built on the information bottleneck principle, it employs a mutual information optimizer to filter irrelevant information across modalities.

SIFF introduces a novel framework for multimodal sentiment analysis by extracting sentiment-specific semantic information from multiple modalities and integrating it with a gated attention fusion module. This module adaptively combines multimodal semantics, mitigating conflicting information while enhancing emotional interaction between modalities.

MIUR introduces an information exchange module that leverages mutual information to promote cross-modal interaction, filter out task-independent noise, and preserve shared task-relevant information. Experiments on MOSI and MOSEI demonstrate significant performance improvements over existing models.

ITHP inspired by neuroscience, refines multimodal representation learning by designating a prime modality and using auxiliary modalities as detectors to streamline information flow. Balancing mutual information minimization and maximization, creates compact latent representations, retaining relevant information while reducing redundancy and enhancing performance.

MIA-NET treats the most emotion-contributing modality as the main modality and employs multimodal interactive attention modules to adaptively select important information from auxiliary modalities, enhancing the main-modal representation while maintaining efficient training with linear computation and stable parameter counts.

As shown in \cref{tab:mosi_transposed} and \cref{tab:mosei_transposed}, our method, MCIB, achieves SOTA performance in F1 score on the CMU-MOSI dataset, surpassing all baseline methods. On the CMU-MOSEI dataset, the corr score of MCIB is only 0.004 behind the best baseline. Overall, compared to all baselines, MCIB demonstrates above-average performance.

The proposed MCIB method does not design task-specific modules for multimodal sentiment recognition. However, experimental results show that it surpasses many baselines and performs closely to the best SOTA results. This highlights the generalizability of our multimodal framework, demonstrating its competitive performance, strong robustness, and adaptability.

\subsection{C. Experimental Details}

\subsubsection{Experimental Settings}

During training, we analyzed the gradients of mutual information and visualized the changes in the loss function. Based on these observations, we adjusted the model parameters, with learning rate and Lagrange multipliers showing the most significant impact on performance. The best-performing configuration was achieved with the following hyperparameters: dropout rate of 0.4, 200 epochs, a batch size of 8, a learning rate of 1e-5, B0\_dim set to 256, and B1\_dim set to 128. Additionally, we employed Optuna for hyperparameter optimization to achieve faster convergence in high-dimensional spaces. The experimental results were robust, with an F1-score 75.64 with an error margin of ± 0.95, demonstrating the model's stability across hyperparameter variations. These methods contributed to reliable and consistent performance across our experiments.

\subsubsection{Implementation Details of IB-based Methods}
To further validate the effectiveness of the proposed MCIB, we compared the experimental results with those of other IB-based strategies. To further validate the effectiveness of the proposed MCIB, we compare the experimental results with those of other IB-based strategies. This section provides the theoretical background and implementation details of the baseline IB-based methods used in the experiments.

Single Information Bottleneck (SIB). In applying the IB principle to multimodal information fusion, a common approach is to use the information bottleneck between each single modality and the target \cite{mai2022multimodal, chen2023multimodal}. The principle, as shown in \cref{sib}, aims to maximize information compression from $x_i$ to $z_i$ while ensuring $z_i$ retains as much information related to the prediction target $y$ as possible.

\begin{equation}
    I(x_i, z_i) - I(z_i, y).
    \label{sib}
\end{equation}

Double Information Bottleneck (DIB). Another strategy is to independently apply the information bottleneck to pairs of modalities (such as text-visual and text-audio), which improves the task relevance and purity of the unimodal representations. The loss function, as defined in \cref{dib} for text-visual pairs \cite{zhang2022information}, enables back-optimization of the representations for each modality ($z_t$, $z_v$, and $z_a$).

\begin{equation}
    L_{tv} = -I_{\theta\psi}(z_t; z_v) + \beta D_{\text{skl}}(p_{\theta}(z_t | X_t) \big|\big| p_{\psi}(z_v | X_v)).
    \label{dib}
\end{equation}

Information-Theoretic Hierarchical Perception (ITHP). The work in \cite{xiao2024neuro} introduces a two-layer information bottleneck. The ITHP method in this work implements a layered architecture for multimodal fusion, creating compact, information-rich hidden states for information flow, as shown in \cref{ithp}. 

\begin{equation}
    I(X_0; B_0) - \beta I(B_0; X_1) + \lambda \left( I(B_0; B_1) - \gamma I(B_1; X_2) \right).
    \label{ithp}
\end{equation}

Although SIB focuses solely on removing target-irrelevant redundancy and DIB only optimizes complementary information between modality pairs, the two-layer ITHP structure restricts information flow to a single direction. In contrast, MCIB simultaneously filters task-relevant information and captures task-relevant complementary features from multiple modalities.

\subsection{D. Shortcut Learning Study}

\setlength{\tabcolsep}{3pt}
\begin{table}[h]
\centering
\begin{tabular}{lccc}
\hline
Method & Precision & Recall & F1-score \\
\hline
MUStARD++ method* & 67.79 & 67.82 & 67.52 \\
MUStARD++ method & 70.07 & 70.03 & 70.03 \\
\hline
TBJE* & 68.94 & 66.94 & 67.04 \\
TBJE & 69.50 & 68.59 & 68.75 \\
\hline
ITHP* & 57.63 & 57.47 & 57.58 \\
ITHP & 68.27 & 68.29 & 68.27 \\
\hline
MCIB* & 75.95 & 75.43 & 75.58 \\
MCIB & 76.14 & 75.83 & 75.64 \\
\hline
\end{tabular}
\caption{
Comparison of cross-dataset results with different training sets. An asterisk (*) indicates training on MUStARD++ with shortcuts and testing on MUStARD++$^{R}$; no suffix means training and testing on MUStARD++$^{R}$.
}
\label{tab:sas}
\end{table}

As shown in Table \ref{tab:sas}, the impact of shortcut learning on multimodal sarcasm detection is demonstrated by training result on different datasets. When trained on the MUStARD++ dataset with shortcuts, the model tends to learn shortcuts rather than the true task intent. In contrast, when trained on MUStARD++$^{R}$, the model cannot rely on shortcuts and must depend on model design and multimodal fusion strategies. When the test set contains no shortcuts (providing an out-of-distribution (OOD) testing environment), models that have only learned shortcuts and lack sufficient information extraction capabilities will experience a decline in performance. As a result, evaluated on the MUStARD++$^{R}$ test set, the results reflect two aspects: the model's dependence on shortcuts and its ability to generalize by learning truly useful information. The results show that our MCIB method neither depends on shortcut learning nor overfits to it, achieving the best fusion strategy. In real-world scenarios, where auxiliary information is less abundant, our method demonstrates superior generalization performance.

\subsection{E. Rethinking and Visualization of Multimodal Fusion}

\begin{figure*}[t]
  \centering
  \includegraphics[width=1\linewidth]{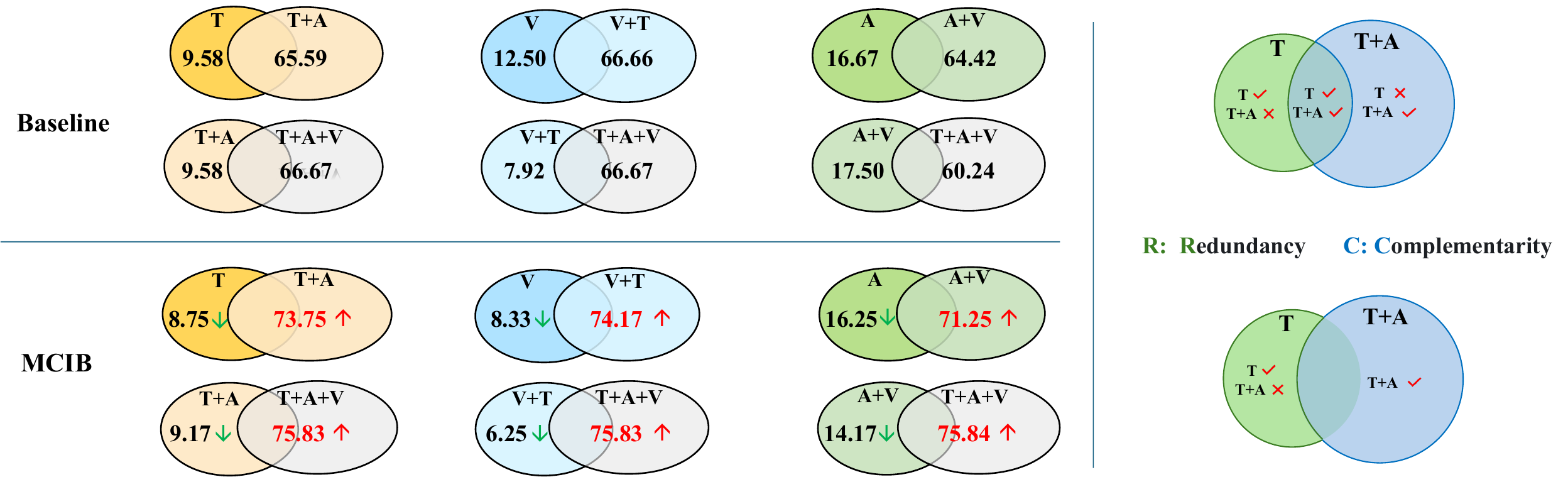}
  \caption{TVA represents the three modalities: Text, Video, and Audio, with the "+" symbol indicating their combinations. The numbers inside the circles represent the proportions: the left circle indicates cases correctly predicted by the left modality but misclassified by the right combination, while the right circle indicates cases correctly predicted by the right combination.}
  \label{figs}
\end{figure*}

\subsubsection{E.1 Rethinking Model Fusion Ineffectiveness}

\setlength{\tabcolsep}{3pt}
\begin{table}[h]
\centering
\begin{tabular}{lccccccc}
\hline
Method & VAT & TV & TA & AV & T & A & V \\
\hline
TBJE   & 68.7 & 63.5 & 62.9 & 68.0 & 66.2 & 68.8 & 66.3 \\
ITHP   & 69.0 & 68.1 & 69.3 & 58.0 & 68.4 & 58.5 & 58.6 \\
MUStARD++ & 71.0 & 68.4 & 69.0 & 65.2 & 68.5 & 66.2 & 65.2 \\
\hline
\end{tabular}
\caption{
Modality ablation results (V: video, T: text, A: audio) on MUStARD++ for TBJE \cite{delbrouck2020transformer}, ITHP \cite{xiao2024neuro}, and MUStARD++ \cite{ray2022multimodal}.
}
\label{tab:1}
\end{table}

We selected three methods for modality ablation experiments. The results in \ref{tab:1} provide quantitative evidence for ineffective modality fusion. The findings show that adding modalities does not always lead to performance improvements; in some cases, it even results in performance degradation (e.g., TBJE: A→AV, V→TV; ITHP: T→TV, TA→VAT; MUStARD++: A→AV, T→TV, etc.). These results highlight the inefficiency in modality fusion in existing methods, which fail to balance the information gain and redundancy introduced by the addition of new modalities.

\subsubsection{E.2 Visualization of Fusion Improvements}

To illustrate the effectiveness of MCIB in removing redundancy and extracting complementarity, we analyze the changes in the proportion of correctly predicted samples on the test set before and after adding modalities. As shown in the \cref{figs}, each Venn diagram depicts the distribution of correctly predicted samples as a percentage of the total test set, highlighting the optimization results when transitioning from one to two, and two to three modalities, for both the MCIB method and baseline approaches.

In \cref{figs} taking TA as an example, the left portion of the diagram shows the proportion of cases correctly predicted by T but misclassified by TA. The middle portion represents cases where both T and TA correctly predict the outcome, while the right portion shows cases correctly predicted by TA but misclassified by T alone. The right portion highlights the redundancy in the A modality, as its addition introduces irrelevant information that interferes with predictions. Conversely, the left portion reflects the complementary information provided by the A modality, as the combination of T and A leads to correct predictions. Analyzing the diagram, reducing the left portion's proportion while increasing the right portion's proportion indicates improved modality fusion. Although the middle and left portions may not simultaneously increase compared to baseline methods, this does not imply a failure to extract complementary information. Our experiments reveal that the middle portion (where both predict correctly) often shows a significant increase compared to the right portion. Moreover, the overall performance after adding modalities substantially improved over baseline methods. Therefore, we use the overall performance of the combined modalities to measure the degree of complementary information utilization.

Compared to the baseline method, our MCIB approach reduces the proportion of the redundancy region \(R\) and increases the proportion of the complementarity region \(C\) across all modality combinations. The experimental results demonstrate the effectiveness of our method in modality fusion, as it minimizes misclassification caused by redundancy while enhancing the extraction and utilization of beneficial information.

\end{document}